%% file: main.tex
\documentclass[10pt,twocolumn,letterpaper]{article}

\usepackage[pagenumbers]{cvpr} 

\input{preamble}

\definecolor{cvprblue}{rgb}{0.21,0.49,0.74}
\usepackage[pagebackref,breaklinks,colorlinks,allcolors=cvprblue]{hyperref}

\newcommand{\sys}{\textsc{VidStamp}\xspace}

\definecolor{todoamber}{RGB}{230,126,34} 

\def\paperID{14464} 
\def\confName{CVPR}
\def\confYear{2026}

\title{\sys: A Temporally-Aware Watermark for Ownership and Integrity in Video Diffusion Models}

\author{Mohammadreza Teymoorianfard\textsuperscript{1}, Siddarth Sitaraman\textsuperscript{2}, Shiqing Ma\textsuperscript{1}, Amir Houmansadr\textsuperscript{1}\\
\textsuperscript{1}University of Massachusetts Amherst, \textsuperscript{2}Brown University\\
{\tt\small mteymoorianf@umass.edu, siddarth\_sitaraman@brown.edu, shiqingma@umass.edu, amir@cs.umass.edu}
}

\begin{document}
\maketitle
\input{sec/abstract}    
\input{sec/intro}

\input{sec/prelims}
\input{sec/threat_model}
\input{sec/methodology}

\input{sec/setup}

\input{sec/results}

\input{sec/conclusion}
{
    \small
    \bibliographystyle{ieeenat_fullname}
    \bibliography{main}
}

\appendix
\input{sec/X_suppl}

\end{document}

%% file: preamble.tex
\usepackage{amsmath}
\usepackage{bbm}

\usepackage{algorithm}
\usepackage{algpseudocode}
\usepackage{enumitem}

\usepackage[skip=0.333\baselineskip]{caption}
\usepackage{booktabs,tabularx,ragged2e}
\usepackage{multirow}
\usepackage{graphicx}
\usepackage[table]{xcolor}

\renewcommand{\arraystretch}{1.3}
\newcolumntype{Y}{>{\hsize=.7\hsize\RaggedRight\arraybackslash}X}
\newcolumntype{a}{>{\columncolor{lightgray}}Y}
\newcolumntype{C}[1]{>{\columncolor{lightgray}}p{#1}} 



%% file: sec/abstract.tex
\begin{abstract}

Video diffusion models can generate realistic and temporally consistent videos. This raises concerns about provenance, ownership, and integrity. Watermarking can help address these issues by embedding metadata directly into the content. To work well, a watermark needs enough capacity for meaningful metadata. It must also stay imperceptible and remain robust to common video manipulations. Existing methods struggle with limited capacity, extra inference cost, or reduced visual quality. We introduce \sys, a watermarking framework that embeds frame-level messages through the decoder of a latent video diffusion model. The decoder is fine-tuned in two stages. The first stage uses static image datasets to encourage spatial message separation. The second stage uses synthesized video sequences to restore temporal consistency. This approach enables high-capacity watermarks with minimal perceptual impact. \sys also supports dynamic watermarking through a control signal that selects message templates during inference. This adds flexibility and creates a second channel for communication. We evaluate \sys on Stable Video Diffusion (I2V), OpenSora, and Wan (T2V). The system embeds 48 bits per frame while preserving visual quality and staying robust to common distortions. Compared with VideoSeal, VideoShield, and RivaGAN, it achieves lower log P-values and stronger detectability. Its frame-wise watermarking design also enables precise temporal tamper localization, with an accuracy of 0.96, which exceeds the VideoShield baseline.
Code: \url{https://github.com/SPIN-UMass/VidStamp}

\end{abstract}


%% file: sec/intro.tex
\section{Introduction}
\label{sec:intro}
The rapid growth of AI-generated video content has created serious challenges for media integrity, security, and public trust. Modern generative models can now produce highly realistic videos from static images or natural language prompts \cite{blattmann2023stable, wang2025wan, opensora, openai2024sora, googledm2025veo3, lumalabs2025dreammachine}. Such generative ability opens new risks of being exploited for misinformation, impersonation, and malicious content editing. These issues highlight the need for reliable methods that can verify model provenance, trace user activity, and detect tampering. In this setting, \emph{watermarking} has emerged as a practical solution for authentication and integrity verification \cite{Abdelnabi2021AWT, Fernandez2023StableSignature, fei2022supervised, zhao2411sok}.

\begin{figure}
    \begin{center}    \centerline{\includegraphics[width=0.42\textwidth]{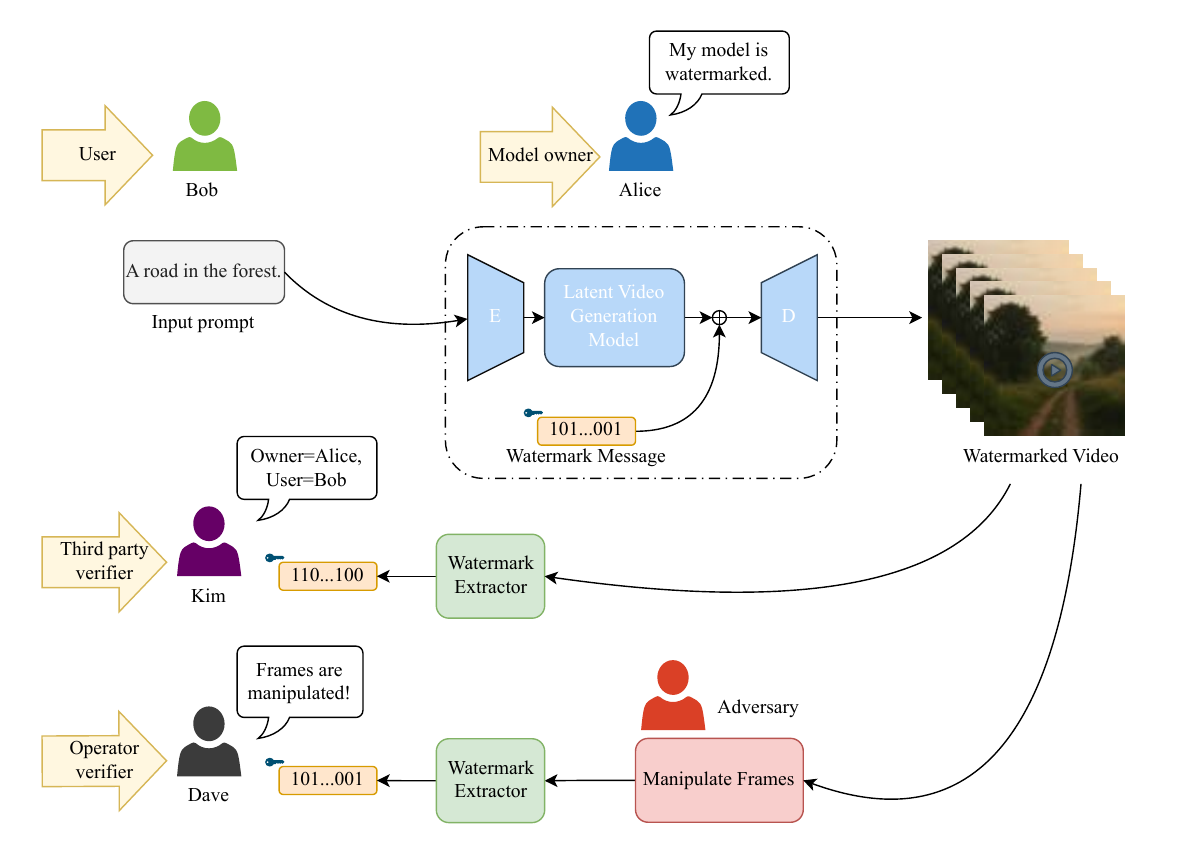}}
    \caption{Overview of the threat model. \sys embeds metadata during generation inside a trusted model environment, while the adversary interacts only with the output video. Two security goals are considered: establishing model ownership and user traceability, and detecting temporal tampering such as frame swaps, insertions, and drops.}
    
    \label{fig:threat_model}
    \end{center}
\end{figure}

As latent video diffusion models become the dominant approach for video generation, their use is expanding rapidly across creative, commercial, and communication platforms. These models can generate high-quality, temporally coherent videos at scale \cite{wang2025lavie, blattmann2023align}. However, when deployed in large production settings, post-processing watermarking introduces a considerable computational cost. Even small inference delays accumulate when millions of videos are produced. For model providers, this cost reduces scalability and long-term efficiency.
The limitation becomes more severe as latent video diffusion models move toward real-time or streaming applications \cite{kodaira2025streamdit}. In these scenarios, even minor post-processing latency can interrupt continuous generation or degrade user experience \cite{henschel2025streamingt2v}. To support such use cases, watermarking must occur inside the generation process rather than after it. Integrated watermarking is therefore essential for scalable and low-latency deployment of modern video diffusion systems.

Watermarking has often been treated as a binary verification problem. If the recovered message exceeds a confidence threshold, the content is considered authentic. This perspective is limited. A watermark can also serve as a communication channel that carries structured metadata \cite{zhao2411sok, jiang2024watermark}. Embedding model identifiers, user information, and digital signatures enables not only authenticity verification but also traceability of the generative model and its users \cite{yang2025gaussian}.

Video is a rich medium for watermarking because each frame can carry its own message, and the sequence of these messages can form a larger payload. Embedding a different message in every frame increases the total capacity and allows more metadata to be stored without affecting visual quality. Frame-wise embedding also supports temporal integrity checks because each frame has a distinct identifier. Any change in the order or content of frames, such as swapping, deletion, or insertion, can be detected by comparing the extracted sequence with the expected one. This combination of high capacity and frame-level traceability makes per-frame watermarking effective for provenance and integrity verification in generative video models.

An effective watermarking framework must balance three key objectives: capacity, robustness, and perceptual quality. Frame-wise embedding naturally supports high capacity, as each frame contributes additional message space without compromising visual fidelity. Robustness ensures that the embedded information can be recovered even after compression, resizing, or other video transformations. Maintaining perceptual quality guarantees that the watermark remains imperceptible while preserving the realism of generated content. These considerations motivate \sys, a temporally-aware watermarking framework that embeds structured, frame-level messages directly within the latent decoding process of video diffusion models.

The idea behind \sys is inspired by Stable Signature \cite{Fernandez2023StableSignature}. That work showed that watermark information can be embedded directly into the generation process by fine-tuning the \emph{decoder of the variational autoencoder (D-VAE)} in a latent diffusion model. This approach removes inference-time overhead and ensures that every generated sample carries a persistent and verifiable signal. \sys follows the same principle but adapts it to latent video diffusion models, which include temporally-aware layers that allow consistent and high-capacity watermark embedding across frames.

\sys uses the temporally-aware D-VAE of modern video diffusion models to embed a sequence of hidden messages at the frame level during generation. In the static watermarking mode, the messages are fixed once the D-VAE is fine-tuned; changing the embedded content would require retraining the model. To address this limitation, \sys also supports a dynamic watermarking mode, where a control sequence is provided to the D-VAE at inference time. This input determines which message template is embedded in each frame, allowing adaptive control of the watermark content without additional training.

A two-stage fine-tuning pipeline is employed. In the first stage, the D-VAE is trained on an COCO \cite{lin2014microsoft} image dataset to learn basic watermarking behavior. In the second stage, the D-VAE is fine-tuned on videos produced by the same diffusion model. This step adapts the model to the temporal structure of video data and promotes consistent message embedding across frames.

The method is applied to three open-source latent video diffusion frameworks: Stable Video Diffusion (SVD) \cite{blattmann2023stable}, OpenSora \cite{opensora}, and Wan \cite{wang2025wan}. \sys achieves strong watermark recovery, high visual fidelity, and robustness against common video distortions. Compared with RivaGAN \cite{zhang2019robust}, VideoSeal \cite{fernandez2024video}, and VideoShield \cite{hu2025videoshield}, \sys achieves better performance in terms of \emph{log P-value}, a statistical metric that jointly accounts for accuracy and capacity.

\noindent\textbf{In summary, this paper makes the following contributions:}

\begin{itemize}
    \item \textbf{A temporally-aware watermarking framework for video diffusion models:}
    \sys embeds structured watermark messages directly within the latent decoding process of video diffusion models. This removes the need for post-processing and avoids any inference-time overhead. Its frame-wise embedding design enables precise temporal tamper localization.
    \item \textbf{A two-stage fine-tuning pipeline:} Our approach first fine-tunes the D-VAE on static images to promote spatial watermark separability, then adapts it to video synthesis to ensure temporal consistency. This training strategy enables \emph{high-capacity watermark embedding} with strong robustness under distortion.
    \item \textbf{Comprehensive evaluation and Superior performance:} \sys is implemented on three open source latent video diffusion models: SVD, OpenSora, and Wan. It achieves high watermark accuracy, strong robustness to common video distortions, and high perceptual fidelity. The method also achieves superior performance to prior baselines (RivaGAN, VideoSeal, VideoShield), while embedding more bits and preserving visual quality.
\end{itemize}

%% file: sec/prelims.tex
\begin{figure*}[t]
    \begin{center}
    \centerline{\includegraphics[width=0.85\textwidth]{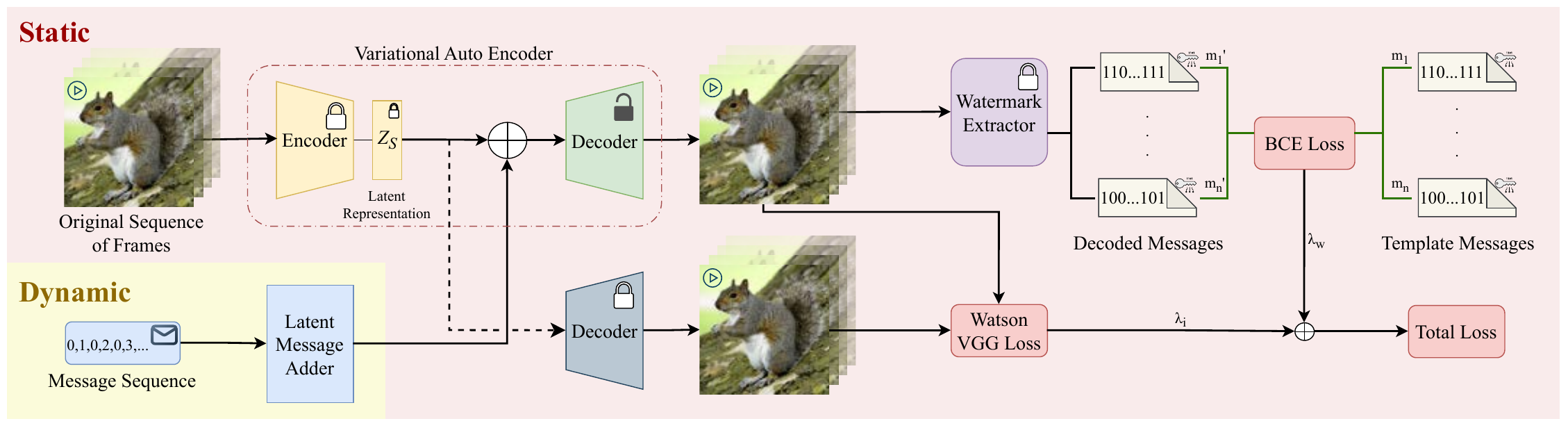}}
    \caption{\sys training pipeline. The decoder is fine-tuned to embed watermark messages into video frames using per-frame message and perceptual losses. A pretrained extractor supervises recovery. The framework supports static watermarking with fixed messages and dynamic watermarking, where a control signal selects message templates during inference.}
 
    \label{fig:MainPipeline}
    \end{center}
\end{figure*}
\section{Related Work}
\label{sec:Preliminaries}

\subsection{Video Diffusion Models and Temporality}
Diffusion models have been extended from images to videos, enabling temporally coherent synthesis. Ho et al.\ extended image denoisers to the spatio-temporal domain using 3D U-Nets \cite{Ho2022VideoDiffusion}. Later work improved scalability through latent video diffusion. He et al.\ \cite{He2022LatentVideo} proposed generating videos in a low-dimensional 3D latent space, reducing computation while supporting long sequences through hierarchical sampling and latent perturbation. Blattmann et al.\ \cite{blattmann2023stable} adapted Stable Diffusion to video by adding temporal layers to the latent model and fine-tuning the decoder with 3D convolutions and temporal attention. These advances enable high-resolution video synthesis with strong temporal consistency and provide the temporal structures that make latent-space watermarking feasible.

\subsection{Watermarking in Generative Models}
Watermarking has been explored across modalities. Early text approaches used lexical substitution, syntactic changes, or stylometry \cite{murphy2007syntax, topkara2006words, atallah2001natural}. More recent methods include statistical token biasing \cite{Kirchenbauer2023Watermark}, neural paraphrasing via the Adversarial Watermarking Transformer \cite{Abdelnabi2021AWT}, and semantic-pattern watermarking such as SimMark \cite{dabiriaghdam2025simmark}.

In images, encoder--decoder systems such as HiDDeN learn imperceptible and robust embeddings \cite{Zhu2018HiDDeN}. A major shift came with model-integrated watermarking in diffusion models. Stable Signature fine-tunes the latent decoder to embed persistent identifiers during generation \cite{Fernandez2023StableSignature}. Tree-Ring Watermarks \cite{Wen2023TreeRing} further enhance robustness by seeding structured signals in the initial noise. These integrated watermarks offer zero inference overhead, strong imperceptibility, and improved resistance to removal compared with post-hoc methods.

\subsection{Video Watermarking Techniques}
Classical video watermarking includes spatial-domain methods that modify pixel values \cite{Hartung1999,Yu2018} and transform-domain methods that embed bits in frequency coefficients to improve robustness \cite{Yu2018,Preda2011}. Many approaches use redundancy and error-correction coding to survive compression, scaling, or desynchronization \cite{Preda2011}. Deep learning methods extend these ideas by training neural encoders and decoders with differentiable distortion layers. For example, Luo et al.\ \cite{luo2023dvmark} distribute payloads across spatial and temporal scales with adversarial losses enforcing invisibility and robustness.

Watermarking for AI-generated video has advanced rapidly. RivaGAN \cite{zhang2019robust} embeds watermarks using an adversarial encoder--decoder but operates only as post-processing. VideoSeal \cite{fernandez2024video} improves robustness through image-then-video training and simulated degradations, including temporal propagation to enhance consistency across frames, yet remains an external module. VideoShield \cite{hu2025videoshield} integrates watermarking into the diffusion sampling process by injecting template noise perturbations and extracting messages via inversion, enabling robust recovery and temporal tamper localization.

Our proposed \sys method embeds watermarks directly in the latent decoding process of the video diffusion model. This integration leverages existing temporal layers to ensure consistency across frames, supports higher capacity and flexible message structures, and avoids post-hoc processing or inversion. As a result, \sys maintains high visual fidelity while enabling reliable provenance, ownership verification, and frame-level tamper detection.

%% file: sec/threat_model.tex
\section{Threat Model}

We consider two distinct security settings for \sys: (1) model ownership and user traceability, and (2) frame integrity and temporal tamper localization. An overview of the threat model is illustrated in Figure~\ref{fig:threat_model}.
In both cases, watermark embedding occurs during generation inside a trusted model environment. The adversary interacts only with the generated videos or the model’s API, never with its internal weights. The watermark extractor is public, allowing third parties to verify provenance without model access.

\subsection{Model Ownership and User Traceability.}

The goal is to establish verifiable ownership of the generative model and to trace each output to its source. During generation, \sys embeds metadata containing model identifiers and user-specific keys directly into the video, allowing later extraction for authentication.

The adversary is a user who tries to claim ownership, misattribute content, or hide the origin of harmful material. They can only access model outputs or API responses and cannot modify internal parameters. Their actions are limited to common video transformations such as resizing, cropping, recompression, brightness or contrast changes, rotation, or mild filtering. Under these conditions, \sys ensures that embedded messages remain recoverable and resistant to typical distortion-based removal attempts.

\subsection{Frame Integrity and Tamper localization}

The second threat scenario addresses content integrity in published or transmitted videos. Here, each frame carries a unique bit sequence that encodes its position or identifier. These per-frame messages allow a verifier to detect and localize temporal edits including frame \emph{insertions, swaps, and drops}.

During verification, extracted messages are matched against the reference sequence; any mismatch or missing key signals a tampering event and indicates its position within the video.

%% file: sec/methodology.tex
\section{Methodology}
\label{sec:Methodology}

\subsection{Overview}

\sys builds on the Stable Signature approach for image diffusion models \cite{Fernandez2023StableSignature}, which showed that fine-tuning the D-VAE can embed a persistent and invisible watermark into generated outputs. 
Building on this idea, \sys applies a similar principle to the video domain, adapting it to the temporal structure of latent video diffusion models. Figure~\ref{fig:MainPipeline} illustrates the overall framework.

In video diffusion architectures, D-VAE reconstructs frames from latent representations, employs 3D convolutional layers and temporal attention mechanisms \cite{li2025diffvsr} that capture correlations across consecutive frames. By adapting this decoder, watermark messages are embedded during generation such that each frame receives a unique bit sequence. These frame-specific identifiers enable per-frame message extraction and facilitate accurate localization of temporal manipulations, including frame insertion, deletion, and swapping \cite{hu2025videoshield}.

Frame-level watermarking is possible only when the D-VAE models temporal structure. Each frame carries a unique message, so the decoder must understand cross-frame relationships to embed distinct patterns consistently. Video models that reuse image decoders lack temporal layers and cannot reliably embed different messages across frames.

For message recovery, \sys uses a pretrained extractor from the HiDDeN framework \cite{Zhu2018HiDDeN}. This network decodes the embedded bit sequences from generated frames and provides supervision during training to ensure accurate watermark reconstruction.

\subsection{Dynamic Watermarking}
In \sys, after fine-tuning, the decoder always embeds a fixed set of messages, and changing them requires retraining the model. This defines the static watermarking mode. To add flexibility at inference, \sys provides a dynamic watermarking mode as well. A control signal is passed to the D-VAE along with the latent representation. This signal contains a sequence of message indices, each mapped to a predefined watermark template. During generation, each frame embeds the message indicated by its index, and the index sequence acts as a second communication channel across time.

The control sequence is injected into the latent space through a lightweight Latent Message Adder (Figure~\ref{fig:MainPipeline}). Each index is embedded, projected to the latent dimensionality, and added to the latent representation before decoding. This allows the decoder to condition on watermark patterns without architectural changes or additional training.

Dynamic watermarking increases flexibility and make inference-time watermark feasible, but introduces trade-offs. The message adder must be adapted for each diffusion model, which reduces portability. Adding the control signal can slightly lower visual quality, and conditioning on multiple templates makes learning harder, reducing effective capacity compared to the static mode.

\subsection{Two Stage Finetuning}

Our training framework, illustrated in the Figure \ref{fig:MainPipeline}, begins with a sequence of frames as the input data, which are passed through a \emph{variational autoencoder (VAE)} to obtain latent representations $Z$. 

Two watermarking modes are supported: \emph{static} and \emph{dynamic}. In the static mode, watermark messages are embedded implicitly by the decoder without any external input. In the dynamic mode, a latent message sequence is injected as an auxiliary signal through a latent message adder, enabling different messages to be assigned to different frames.

Depending on the selected mode, the latent message signal may be added to the latent representation of input frames before decoding. The resulting latent features are then processed by a temporally-aware VAE decoder. This decoder generates video frames that contain imperceptible embedded messages. Each frame is associated with a fixed bit sequence, and a pretrained message extractor is employed to recover the embedded bits from the generated frames. The recovery process is supervised using a binary cross-entropy (BCE) loss:

\begin{equation}
\mathcal{L}_{\text{msg}} = -\frac{1}{N} \sum_{i=1}^{N} \left[y_i \log(\hat{y}_i) + (1 - y_i)\log(1 - \hat{y}_i)\right]
\end{equation}

\noindent where $y_i$ and $\hat{y}_i$ denote the ground truth and predicted bits, respectively. In parallel, we compute a Watson-VGG perceptual loss \cite{czolbe2020loss} on a frame-by-frame basis to maintain the semantic integrity and visual quality of the generated outputs. This perceptual loss, denoted $\mathcal{L}_{\text{perc}}$, compares deep feature activations between the original and reconstructed frames, emphasizing human-perceptible discrepancies. The final training objective is a weighted sum of the message loss and perceptual loss:
\begin{equation}
\mathcal{L}_{\text{total}} = \lambda_w \cdot \mathcal{L}_{\text{msg}} + \lambda_i \cdot \mathcal{L}_{\text{perc}}
\end{equation}

\noindent where $\lambda_w$ and $\lambda_i$ are tunable hyperparameters that control the trade-off between bit accuracy and visual quality. During training, only the decoder is updated, while the encoder and message extractor remain frozen. This setup enables the decoder to learn effective message embedding strategies without compromising on generation fidelity.

Our training pipeline uses a two-stage fine-tuning process to enable the decoder to embed robust and temporally consistent watermarks without degrading video quality. In the first stage, the decoder is trained on COCO images \cite{lin2014microsoft}, where each batch of independent images is treated as a pseudo-video. Each image acts as a separate frame, which encourages the decoder to treat frames independently and to associate different spatial features with different message embeddings. This step initializes the decoder’s ability to separate and encode frame-specific watermarks.


However, training only on images exposes the decoder to static content and does not teach it the temporal statistics present in videos. To address this, the second stage fine-tunes the decoder on videos synthesized by the same diffusion model. This exposes the decoder to frame-to-frame correlations and motion continuity, improving perceptual coherence and stabilizing message embedding across time. Together, the two stages balance spatial watermark capacity with temporal consistency.

\subsection{Temporal Tamper Localization}
Temporal tampering is detected by comparing the frame-wise watermark messages of a test video with the reference message sequence used during generation. The procedure follows the same high-level idea as the verification algorithm in VideoShield \cite{hu2025videoshield}, but in \sys each frame contains a unique embedded message that serves as its key for localization.

During verification, each extracted message is compared against all reference keys using Hamming similarity. A frame is marked as inserted if its similarity to every reference key falls below a defined threshold; otherwise, it is assigned to the closest match. The predicted frame order is then aligned with the reference sequence to identify and localize insertions, deletions, or swaps. The full description is provided in Algorithm~\ref{alg:tamper} in Appendix.

%% file: sec/setup.tex
\section{Experimental Setup}
\subsection{Models and Baseline Methods}

For our experiments, we evaluate \sys on three open-source latent video diffusion models: Stable Video Diffusion (SVD) \cite{blattmann2023stable}, an image-to-video model, and OpenSora \cite{opensora,openai2024sora} and Wan \cite{wang2025wan}, both text-to-video models. Each model can generate temporally coherent videos from an image or text prompt, and we embed a 48-bit message into every frame. We compare \sys against three representative watermarking baselines: RivaGAN \cite{zhang2019robust} and VideoSeal \cite{fernandez2024video}, which embed watermarks after generation, and VideoShield \cite{hu2025videoshield}, which integrates watermarking during sampling. Implementation details for all models and baselines are provided in Appendix \ref{subsec:modelconf}.

\subsection{Metrics}
\noindent\textbf{Bit Accuracy.} Bit accuracy measures the proportion of correctly extracted bits from the watermarked video relative to the original message. Given the original bitstring $m \in \{0,1\}^n$ and the extracted bitstring $\hat{m}_i$, the bit accuracy is:
\begin{equation}
    \text{Bit Accuracy} = \frac{1}{n} \sum_{i=1}^{n} \mathbbm{1}[m_i = \hat{m}_i]
\end{equation}      
\textbf{Log P-Value.} Following VideoSeal \cite{fernandez2024video}, we also report the log P-value, which better reflects the statistical confidence in watermark detectability. Unlike bit accuracy, the log P-value accounts for the probability of false positives, making it a more suitable metric when comparing models with different watermark capacities. Given a watermark length of $L$, and an observed bit accuracy $a$:
        \begin{equation}
        \log P = \log_{10} \left( \sum_{k=\lceil aL \rceil}^{L} \binom{L}{k} (0.5)^L \right)
        \end{equation}

    This metric corresponds to the probability of achieving the observed bit accuracy or higher by random guessing. A lower log P-value indicates a stronger, more detectable watermark. We use this metric as our primary benchmark for watermark evaluation, as it normalizes across varying capacities and bit lengths.

  \noindent\textbf{Video Quality Metrics.} To evaluate whether watermark embedding affects the perceptual quality of the generated videos, we adopt a suite of standard video quality metrics inspired by VideoShield \cite{hu2025videoshield} and measured using the VBench \cite{huang2024vbench} evaluation toolkit: \textit{Subject Consistency, Background Consistency, Motion Smoothness, Aesthetic Quality, and Imaging Quality} described in detail in Appendix~\ref{subsec:videoqual}.

\subsection{Datasets}


We use two datasets in our training pipeline, corresponding to the two-stage fine-tuning process in Section~\ref{sec:Methodology}. The first stage uses an COCO \cite{lin2014microsoft} image dataset to learn spatially distinct message embeddings. The second stage uses a set of videos generated by the same diffusion model, based on a video-prompt dataset, to adapt the decoder to temporal statistics. All details on dataset construction, prompt selection, and video generation are provided in Appendix~\ref{subsec:datasetdetails}.

%% file: sec/results.tex
\section{Experimental Results}

\label{sec:Experiments}

\subsection{Performance of \textbf{\sys}}
\label{sec:performance_vidstamp}

\begin{table}[ht]
\centering
\caption{Static watermarking performance across three latent video diffusion models. \sys achieves consistently high bit accuracy and strong robustness across SVD, OpenSora, and Wan, while preserving perceptual quality. The total embedded payload reflects each model's number of frames and per-frame capacity.}
\label{tab:static_watermark}
\setlength{\tabcolsep}{1pt}
\resizebox{0.45\textwidth}{!}{
\begin{tabular}{l|ccccccccc}
\toprule
{Model} & Bit Acc $\uparrow$  & Length $\uparrow$  & log(p) $\downarrow$  & Avg Robust Acc $\uparrow$ & Avg Quality $\uparrow$ \\
\midrule
SVD & 0.950 & 768 & -166.65 & 0.769 &  0.836 \\ 
OpenSora & 0.983 & 624 & -166.48 & 0.794 & 0.747\\ 
Wan  & 0.933 & 624 & -123.32 & 0.753 & 0.645 \\ 
\bottomrule
\end{tabular}
}
\end{table}

\paragraph{Overall Performance on Three Models.}
Table~\ref{tab:static_watermark} presents the static watermarking results for \sys across three latent video diffusion models: SVD, OpenSora, and Wan. \sys achieves strong bit accuracy and stable robustness on all three systems, demonstrating that the static watermarking framework generalizes well across architectures with different temporal structures and latent spaces. However, the perceptual quality scores in Table~\ref{tab:static_watermark} should not be compared directly. The models operate at different native resolutions and rely on different sampling settings, and both OpenSora and Wan normally run at much higher resolutions than those used in our evaluation. Due to computational constraints, they were tested under reduced configurations, which limits the attainable image quality. As a result, cross-model quality comparisons are not meaningful, even though the watermarking accuracy remains consistent across all models.

\paragraph{Dynamic Watermarking Results on SVD.}
Table~\ref{tab:dynamic_watermmark} reports the dynamic watermarking results on the SVD model. When the number of available template messages is small, dynamic watermarking achieves high bit accuracy and produces videos with quality close to the static baseline. As the number of template messages increases, the task becomes more difficult for the decoder, since it must distinguish between a larger set of message patterns while preserving the visual content. This leads to a drop in bit accuracy. When the task is well learned, a slight reduction in quality is expected as a natural result of watermark embedding. However, when the task becomes too difficult and the decoder fails to learn the message patterns, the quality remains similar, since the model no longer responds to the injected control signal. These observations illustrate the balance between control flexibility and reliable decoding in the dynamic setting.

\paragraph{Comparison Between Static and Dynamic Modes.}
Dynamic watermarking adds controllability at inference time and creates an additional communication channel through its control signals. However, injecting the latent control input reduces visual quality more than the static mode and lowers effective capacity, since the decoder must learn a harder and less stable task. Dynamic watermarking is evaluated on SVD as a representative model, since the message-adder must be adapted to each architecture and the goal here is to demonstrate feasibility rather than provide full cross-model benchmarking. Overall, dynamic watermarking is a useful extension but offers lower accuracy and perceptual fidelity than the static configuration.

\begin{table*}[ht]
\centering
\caption{Dynamic watermarking performance on SVD. Results are shown for both watermarking channels. First row shows the non-watermarked model included for reference.}
\label{tab:dynamic_watermmark}
\resizebox{0.75\textwidth}{!}{

\begin{tabular}{c|c|ccc|ccc}
\toprule

\multirow{2}{*}{{Number of Messages}}&\multirow{2}{*}{{Avg Quality $\uparrow$} } & \multicolumn{3}{c|}{{Channel 1}} & \multicolumn{3}{c}{{Channel 2}} \\

&& Bit Acc $\uparrow$  & Length $\uparrow$  & Avg Robust Acc $\uparrow$  & Bit Acc $\uparrow$  & Length $\uparrow$  & Avg Robust Acc $\uparrow$ \\
\midrule
W/O & 0.715 & - & - & - & - & - & - \\

\cline{1-8}

2 & 0.685 & 0.999 & $2 \times 16$ & 0.832 & 0.999 & $1 \times 16$ & 0.967 \\
4 & 0.671 & 0.983 & $4 \times 16$ & 0.829 & 0.988 & $2 \times 16$ & 0.903 \\
8 & 0.683 & 0.732 & $8 \times 16$ & 0.655 & 0.421 & $3 \times 16$ & 0.364 \\
16 & 0.696 & 0.557 & $16 \times 16$ & 0.530 & 0.075 & $4 \times 16$ & 0.075 \\

\bottomrule
\end{tabular}
}
\end{table*}

\subsection{Comparison with Baselines}
\label{sec:baseline_comparison}

\begin{table*}[ht]
\centering
\small
\caption{Comparison of watermarking methods. \sys embeds 768 bits per video (48 bits/frame × 16 frames), offering higher capacity than prior work. The table reports bit accuracy, log P-value (lower is better), and five VBench-based quality metrics. The first row reports the output quality of the underlying SVD model without watermarking, serving as a perceptual upper bound.}
\renewcommand{\arraystretch}{1.3}
\label{tab:baseline_comparison}
\centering
\resizebox{0.85\textwidth}{!}{
\setlength\tabcolsep{2pt}
\begin{tabular}{c|c c c | c c c c c c | c}
    \toprule
    
    {Method} &\multirow{2}{*}{Length $\uparrow$} & \multirow{2}{*}{Bit Acc  $\uparrow$} & \cellcolor{gray!30} &\multicolumn{7}{c}{Video Quality $\uparrow$}\\

      & & & \cellcolor{gray!30}\multirow{-2}*{$\log_{10}(p)$ $\downarrow$} & Subject consistency &
                        Background consistency &
                        Motion smoothness &
                        Aesthetic quality &
                        Imaging quality &&
                        \cellcolor{gray!30}Avg 
                        \\
    
    \midrule
    
    W/O& $-$ & $-$ & \cellcolor{gray!30}$-$ & 0.961 & 0.957 & 0.964 & 0.610 & 0.697 && \cellcolor{gray!30}0.838 \\

    \cline{1-11}
    
    VideoShield& 512 & 0.995 &  \cellcolor{gray!30}{-149.0} & 0.964 & 0.960 &0.963 &0.587 &0.699 && \cellcolor{gray!30}{0.835} \\
    
    VideoSeal& 96 & 0.979 & \cellcolor{gray!30}{-26.9} & 0.961 &0.956 &0.967 &0.561 &0.672 && \cellcolor{gray!30}{0.823} \\

    RivaGan& 32 & 0.970 & \cellcolor{gray!30}{-9.6} & 0.960 & 0.953 &0.964 &0.598 &0.690 && \cellcolor{gray!30}{0.833} \\

    \textbf{\sys}& 768 & 0.950 & \cellcolor{gray!30}{\textbf{-166.65}} & 0.959 & 0.955 & 0.961 & 0.606 & 0.699 &&\cellcolor{gray!30}{\textbf{0.836}} \\
    \bottomrule
    \end{tabular}
}
\end{table*}

\paragraph{Video Quality Preservation.}
Table~\ref{tab:baseline_comparison} compares \sys with prior watermarking approaches across video quality metrics. 
Among all methods, \sys achieves the strongest balance between watermark embedding and perceptual fidelity. 
Its average quality score of 0.836 is almost identical to the unwatermarked SVD output (0.838), which shows that watermark embedding inside the decoder preserves spatial detail and temporal coherence. 
In both Aesthetic Quality and Imaging Quality, \sys matches or surpasses competing methods. 
These results confirm that decoder-level embedding causes less visible degradation than post-hoc neural encoders such as VideoSeal or RivaGAN.

\paragraph{Embedding Capacity and Watermark Detectability.}
\sys also offers the highest embedding capacity, inserting 768 bits per video (48 bits per frame), much higher than VideoSeal (96 bits), RivaGAN (32 bits), or VideoShield (512 bits). 
Although VideoShield and VideoSeal report slightly higher raw bit accuracy, bit accuracy alone does not reflect the length of the embedded message and therefore does not allow a fair comparison across methods. 
To address this, we follow prior work and report \( \log P \), which capture the statistical confidence of recovering the correct message while accounting for both capacity and accuracy. 
\sys achieves the lowest log P-value ($-166.65$) among all baselines, indicating the strongest watermark detectability even at significantly higher capacity. 
This result shows that decoder-based watermark integration yields a more reliable and statistically verifiable watermark than both post-hoc and sampling-based alternatives.

\subsection{Robustness}
\label{sec:robustness}

\begin{figure}
    \begin{center}    \centerline{\includegraphics[width=0.4\textwidth]{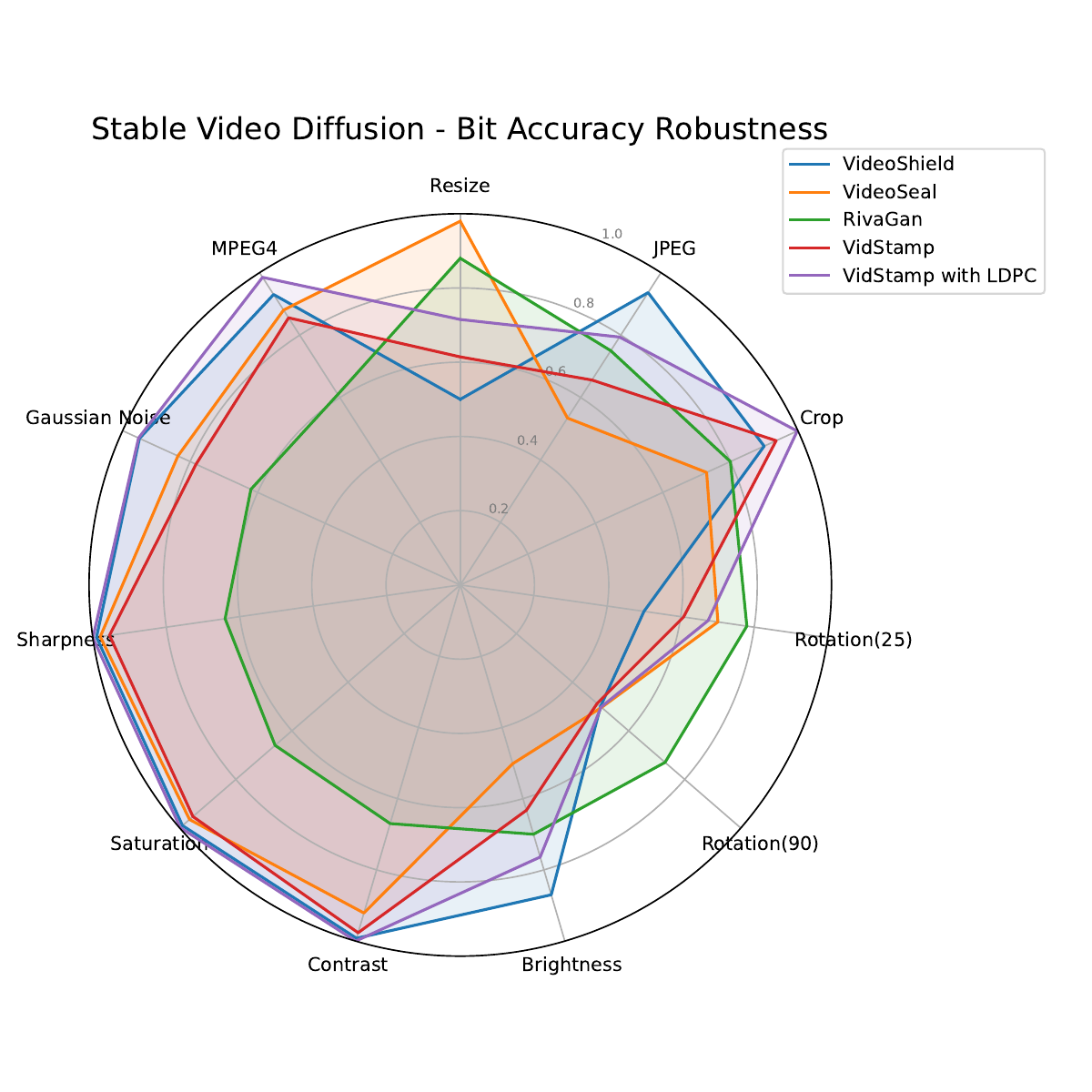}}
    \caption{Bit accuracy under 11 video distortions. \sys maintains high accuracy across most distortions, and the LDPC variant improves performance.}

    \label{fig:robustness_acc}
    \end{center}
\end{figure}

\begin{table*}[ht]
\centering
\small
\caption{\( \log P \) comparison under 11 common video distortions. Lower values indicate stronger watermark detectability. \sys achieves the best overall robustness across distortions.}

\renewcommand{\arraystretch}{1.3}
\label{tab:robustness_log}
\centering
\resizebox{0.9\textwidth}{!}{
\setlength\tabcolsep{2pt}
\begin{tabular}{c|ccccccccccccccccccccc} 
    \toprule

    Method & Resize & JPEG  & Crop & Rotation(25) & Rotation(90) & Brightness & Contrast & Saturation & Sharpness & Gaussian Noise & MPEG4 \\
    
    \midrule
    VideoShield& -0.29& {\textbf{-103.25}}& -83.21& -0.29& -0.29& {\textbf{-69.88}}& -144.67& -142.66& -142.66& {\textbf{-111.82}}& {\textbf{-99.80}} \\
    
    VideoSeal& {\textbf{-26.91}}& -0.62& -5.84& -4.56& -0.34& -0.34& -18.79& -23.73& -26.91& -11.70& -14.89 \\
    
    RivaGan& -5.89& -2.46& -3.57& -2.98& {\textbf{-2.46}}& -1.99& -1.60& -1.60& -1.26& -0.97& -0.97 \\

    \textbf{\sys with LDPC} & -12.05 & -22.03 & -77.06 & -8.01 & -0.32 & -17.60 & -77.06 & -77.06 & -77.06 & -56.93 & -70.61\\

    \textbf{\sys}& -9.95& -17.76& {\textbf{-153.24}}& {\textbf{-8.94}}& -0.12& -13.06& {\textbf{-195.14}}& {\textbf{-169.23}}& {\textbf{-171.86}}& -58.41& -94.71 \\
    \bottomrule
    \end{tabular}
}
\end{table*}

\sys embeds 48 bits in each frame, which provides more capacity than prior work. Part of this capacity can be used for forward error correction. We apply a low-density parity-check (LDPC) code \cite{gallager1962low, richardson2001design} to each frame, where 16 bits store the message and 32 bits store parity. This improves bit accuracy under distortion by correcting extraction errors. We also evaluate word accuracy, where a word is considered correct only if all 16 data bits are recovered without error. As shown in Appendix~\ref{subsec:word_accuracy}, LDPC leads to even larger gains in this setting: word accuracy rises to 0.95 or higher in the clean case and under several distortions,

To evaluate robustness, we apply eleven common video distortions used in prior work. 
The full description of these distortions is available in Appendix~\ref{sec:appendix_distortions}. 
Figure~\ref{fig:robustness_acc} shows that \sys performs on par with the best baselines for most transformations, even though it embeds more bits per frame. 
This indicates that decoder based watermarking maintains resilience under both geometric and photometric changes. 
Figure~\ref{fig:robustness_acc} also shows that LDPC improves bit accuracy for many distortions, confirming that some of the extra capacity can be used to increase robustness when needed.

To compare methods with different capacities, we follow VideoSeal and report log P-values in Table~\ref{tab:robustness_log}. 
This metric reflects statistical confidence in message recovery. 
\sys achieves the best overall log P-value across methods, with the lowest score in several challenging distortions such as crop, contrast, rotation, saturation, and sharpness. 
LDPC does not improve log P-values because it reduces the net payload, and the metric depends on both capacity and accuracy. 
This creates a trade-off: LDPC increases robustness but lowers the effective message length.

\subsection{Tamper Localization}
\label{sec:tamper_localization}

\sys enables detection and localization of temporal manipulations such as frame swaps, insertions, and deletions. The detailed setup and attack configurations are described in Appendix~\ref{sec:appendix_tamper}. During verification, each extracted frame message is compared with reference keys using Hamming similarity. Frames whose similarity falls below a defined threshold are flagged as inserted, while others are aligned with the closest matching reference key. The impact of different threshold values is analyzed in Appendix ~\ref{sec:appendix_tamper}.

Table~\ref{tab:tamper_localization} summarizes localization accuracy under different tampering scenarios. \sys achieves an average accuracy of \emph{0.96} across all attack types, substantially outperforming the VideoShield baseline (\emph{0.936}). These results confirm that the per-frame watermarking design effectively preserves temporal identifiers, enabling precise localization of tampered regions within video sequences.

\begin{table}[t]
\centering
\caption{{Frame-level tamper localization accuracy.} 
\sys achieves higher accuracy than VideoShield across all attacks.}
\label{tab:tamper_localization}
\resizebox{0.45\textwidth}{!}{
\setlength{\tabcolsep}{5pt}
\begin{tabular}{l|cccc}
\toprule
{Method} & Frame Swap & Frame Insert & Frame Drop & Average \\
\midrule
 VideoShield & 0.935 & 0.937 & 0.936 & 0.936\\
 \textbf{\sys} & \textbf{0.959} & \textbf{0.962} & \textbf{0.959 }& \textbf{0.960} \\
\bottomrule
\end{tabular}
}
\end{table}

\subsection{Insights from Ablation Study}
\label{sec:ablation_insights}

The ablation studies provide several insights into the design of \sys.  
First, as shown in Appendix~\ref{sec:appendix_cost}, integrating watermarking into the generation process removes the inference-time overhead seen in post-hoc methods such as VideoSeal, which require an extra neural pass for every video. Although \sys requires a short fine-tuning step, this cost is paid only once, and it adds no overhead during inference, making model-integrated embedding well suited for large-scale deployments. 

Second, Appendix~\ref{sec:appendix_secondstage} shows that the second stage of decoder fine-tuning is critical for achieving robustness. Training on videos adapts the model to temporal correlations, improving bit accuracy from 0.70 to 0.98 on clean data and significantly boosting performance under distortions. This stage also stabilizes frame-wise watermark extraction, leading to more consistent results across time.  

Finally, the tamper-localization analysis in Appendix~\ref{sec:appendix_tamper} and ~\ref{sec:appendix_tamper_agressive} demonstrates that \sys accurately detects and localizes frame insertions, deletions, and swaps. A similarity threshold of $\tau=0.8$ yields the best trade-off between precision and recall, achieving over 95\% accuracy even under combined attacks. These findings confirm that the per-frame watermarking design provides both efficiency and robustness for practical deployment.

%% file: sec/conclusion.tex
\section{Discussion and Conclusion}
\label{sec:Conclusion}

\sys also has several limitations. The method uses a two-stage training pipeline, which requires some additional time and compute beyond the base model. Also, our approach assumes full access to the video diffusion model, including the ability to modify the decoder. This limits deployment in closed or proprietary systems. In addition, while \sys is robust to common video corruptions, it has not been tested against targeted removal attacks. These stronger adversaries may actively attempt to erase or weaken the watermark, and evaluating such scenarios is an important direction for future work.

In conclusion, \sys provides an effective framework for embedding high-capacity, frame-level watermarks directly into video diffusion models. It preserves perceptual quality, introduces no inference-time cost, and supports strong temporal tamper localization. Our results show that decoder-integrated watermarking is both practical and robust for modern generative video systems. We hope this work encourages further research on secure watermarking, adversarial robustness, and reliable provenance for AI-generated video.

%% file: sec/X_suppl.tex
\clearpage

\input{sec/appendix_setup}
\input{sec/appendix_experimental}
\input{sec/abalation}

%% file: sec/appendix_setup.tex
\section{Experimental Setup Details}
\label{sec:app_setup}

\subsection{Model Configuration Details}
\label{subsec:modelconf}
For SVD, we generate 16 frames per video. While the decoder is fine-tuned at a spatial resolution of 256×256, inference is performed at 512×512 resolution to evaluate robustness and generalization under higher-fidelity outputs. Both OpenSora 2 and WAN 2.1 generate 13 frames per video, with a spatial resolution of 256×256. In our main experiments, we embed fixed-length 48-bit messages into each of the 16 frames, resulting in a total payload of 768 bits per SVD video and 624 bits per OpenSora 2 and WAN2.1 video.

\subsection{Video Quality Metrics}
\label{subsec:videoqual}
To evaluate whether watermark embedding affects the perceptual quality of the generated videos, we adopt a suite of standard video quality metrics inspired by VideoShield \cite{hu2025videoshield} and measured using the VBench \cite{huang2024vbench} evaluation toolkit:
\begin{itemize}
\item \textit{Subject Consistency:} Measures whether the appearance of key subjects (e.g., people, animals, objects) remains consistent throughout the video. This is computed using DINO \cite{caron2021emerging} feature similarity across frames. Higher values indicate less visual drift and better temporal identity preservation.
 \item \textit{Background Consistency:} Evaluates the temporal coherence of background scenes by comparing CLIP-based \cite{radford2021learning} features across frames. This captures whether the background remains stable and realistic over time, without sudden changes or artifacts.
\item \textit{Motion Smoothness:} Assesses how physically plausible and continuous the motion is across frames. VBench uses priors from a video frame interpolation model \cite{li2023amt} to evaluate whether the motion adheres to real-world dynamics. Lower jitter and abrupt movement yield higher scores.
\item \textit{Aesthetic Quality:} Reflects the visual appeal of individual video frames. It is computed using the LAION aesthetic predictor \cite{laion2022aesthetic}, which considers aspects such as color harmony, composition, and artistic quality. Higher scores correspond to more aesthetically pleasing frames.
\item \textit{Imaging Quality:} Measures technical image fidelity, such as absence of noise, blur, or artifacts. This is evaluated using the MUSIQ \cite{ke2021musiq} model trained on the SPAQ \cite{fang2020perceptual} dataset. It captures whether the video frames resemble high-quality photographs in terms of clarity and exposure.
\end{itemize}

\subsection{Dataset and Preprocessing Details}
\label{subsec:datasetdetails}

For the first stage of decoder fine-tuning, we use the COCO dataset~\cite{lin2014microsoft}. Each image is treated as an independent frame within a pseudo-video batch to encourage spatially distinct message embeddings without relying on temporal cues. 

For the second stage, we use the VBench prompt set~\cite{huang2024vbench}, which contains 800 prompts spanning eight semantic categories. To produce training and evaluation videos, we generate conditioning images for each prompt using Stable Diffusion 2.1~\cite{Rombach_2022_CVPR}. Of the 800 prompts, 640 (80\%) are used for generating videos for the second-stage fine-tuning, while the remaining 160 (20\%) are reserved for evaluation. These evaluation videos are used to measure watermark accuracy, log P-value, robustness to distortions, video quality, and comparative performance against baseline watermarking methods.

\subsection{Distortion Attacks Configuration}
\label{sec:appendix_distortions}
We consider an adversary that is capable of applying eleven common video distortions to the generated videos. In particular, the distortions we tested are the following:
\begin{itemize}
    \item \textit{Resize:} Resizing the video with the resulting video $70\%$ of its original dimensions.
    \item \textit{JPEG:} Encoding all frames using JPEG compression with quality parameter 50. 
    \item \textit{Crop:} Cropping the borders so that the resulting video has $70\%$ of its original dimensions.
    \item \textit{Rotation(25):} Rotating the video by $25^{\circ}$.
    \item \textit{Rotation(90):} Rotating the video by $90^{\circ}$.
    \item \textit{Brightness:} Increasing the brightness by a factor of 2.
    \item \textit{Contrast:} Increasing the contrast by a factor of 2.
    \item \textit{Saturation:} Increasing the saturation by a factor of 2.
    \item \textit{Sharpness:} Increasing the sharpness by a factor of 2.
    \item \textit{Gaussian Noise:} Adding gaussian noise with mean of zero and standard deviation of 0.1.
    \item \textit{MPEG4:} Encoding all frames using MPEG-4 compression.
\end{itemize}
Note that the Resize, Crop, Rotate, Brighten, Contrast, Saturation, and Sharpness transformations are done using the PyTorchvision library \cite{torchvision2016}.

%% file: sec/appendix_experimental.tex
\section{Experimental Results}
\label{sec:app_experimental}
\subsection{Word accuracy with LDPC Codes}
\label{subsec:word_accuracy}
\begin{figure*}    
    \begin{center}    \centerline{\includegraphics[width=1\textwidth]{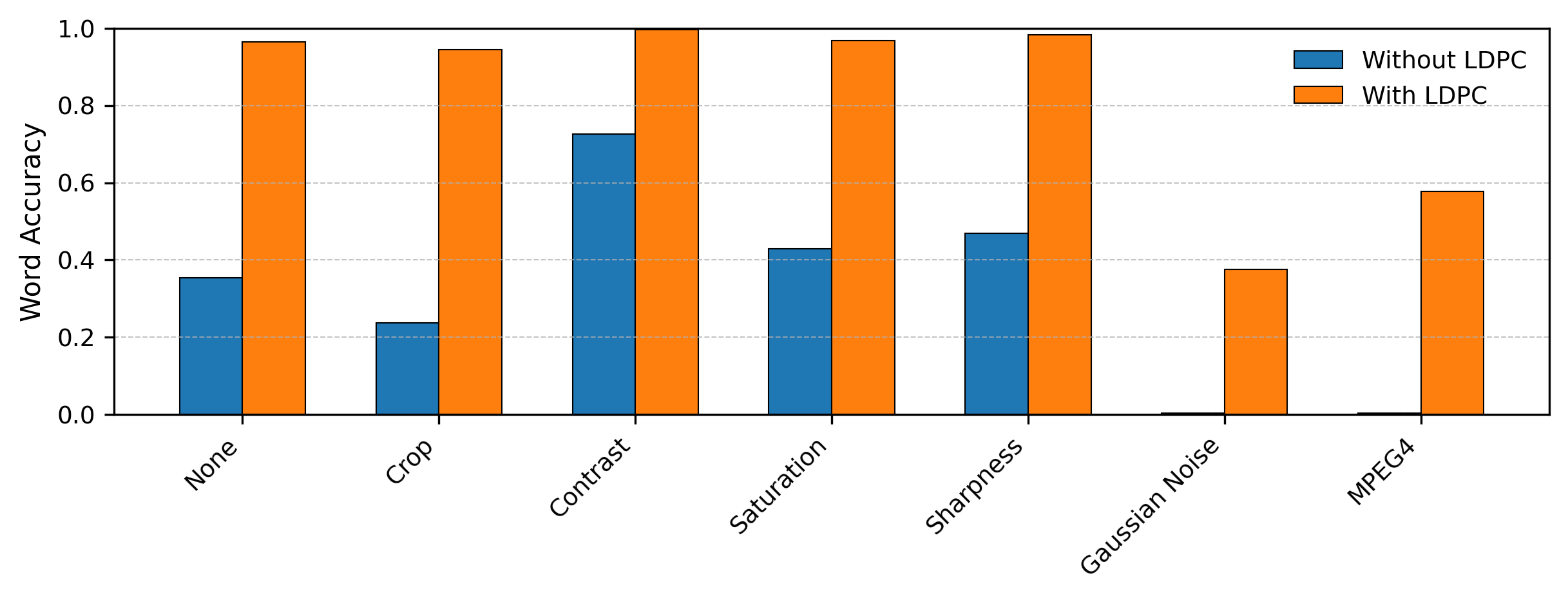}}
    \caption{LDPC codes improve word accuracy significantly when there is no explicit distortion applied ("none") and when there are other common distortions .}
    \label{fig:ldpc}
    \end{center}
\end{figure*}
We use LDPC codes with 16 data bits and 32 parity bits to make up a single 48-bit word, i.e., the data rate is 33.3\%. Word accuracy is the fraction of times the 16 data bits in the word were recovered without errors. The watermark is a sequence of words and can be more reliably recovered when the word accuracy is higher. 
Figure~\ref{fig:ldpc} shows that word accuracy improves drastically from 0.35 to 0.96 when LDPC codes are applied for the common case when there was no distortion beyond our watermarking process. For the crop,   saturation and, sharpness distortions the word accuracy improves to more than 0.95 in each case, a doubling or more of the accuracy. However, LDPC codes did not significantly improve the word accuracy for resizing, jpeg, rotation, and brightness operations. 

%% file: sec/abalation.tex
\section{Ablation Study}

\subsection{Computational Cost Comparison}
\label{sec:appendix_cost}

Recent post-hoc watermarking methods, such as \textbf{VideoSeal}~\cite{fernandez2024video}, embed watermarks by running an additional neural network on the generated video after decoding. While effective in controlled benchmarks, these approaches introduce a substantial computational overhead when scaled to production environments. The public implementation of VideoSeal reports an embedding cost of approximately \textbf{42~GFLOPs} for a 5-second, $256\times256$ video clip. Although this cost may appear moderate for individual videos, it becomes significant at large scale.

\paragraph{Cost of our method.}
In contrast, our method requires only a single fine-tuning stage and adds \textbf{no computational cost at inference}. The training procedure to integrate the watermarking mechanism into the model takes approximately \textbf{20 minutes on a single NVIDIA A100}. Once training is completed, watermark embedding becomes a byproduct of the generative process itself. No auxiliary network is invoked, and the generation-time compute cost remains identical to that of the base video model.

\paragraph{Impact on large-scale text-to-video platforms.}
Consider Google's Veo~3~\cite{googledm2025veo3}, which has already generated over 70 million videos within the first few months after release. Even if all clips were processed at VideoSeal's baseline resolution, the total embedding cost would exceed \textbf{$3\times10^{18}$ FLOPs}. In practice, however, Veo typically produces 8-second videos at \textbf{720p–1080p} resolution. Assuming that computational cost scales linearly with duration and quadratically with spatial resolution, the per-video cost increases by approximately \textbf{$\times$28.5}, yielding around \textbf{1.2 TFLOPs} per 8-second 1080p clip. For the current corpus size, this corresponds to roughly \textbf{$8\times10^{19}$ FLOPs} of additional computation—about \textbf{75 GPU-hours} on a 300~TFLOP/s accelerator. Such overhead directly increases latency, scheduling complexity, and I/O demand in large-scale video generation pipelines.

\paragraph{Violation of real-time constraints.}
Even at 42~GFLOPs per 5~s / $256^2$ clip, the per-frame compute exceeds the latency budget of most real-time pipelines. At 1080p, the estimated \textbf{1.2~TFLOPs per 8~s} corresponds to roughly \textbf{150~GFLOPs/s} of sustained additional computation, which is well beyond the typical budget allocated for auxiliary processing in real-time or streaming generation systems.

\paragraph{Discussion.}
These estimates highlight a fundamental limitation of post-hoc watermarking: each additional inference pass introduces significant computational overhead and latency. In contrast, \sys requires only a \textbf{one-time training cost} and incurs \textbf{zero inference-time overhead}. This makes it far more practical for industrial-scale deployment, where even small per-video costs become prohibitive when multiplied across tens of millions of generations.

\subsection{Contribution of Second Stage Fine-tuning}
\label{sec:appendix_secondstage}

\begin{table}[t]
\centering
\caption{{Impact of second-stage fine-tuning on robustness.} 
Bit accuracy (\%) of the decoder after the first (image-only) and second (video-adapted) fine-tuning stages under common video distortions. The second stage significantly improves robustness across nearly all transformations.}
\label{tab:second_stage}
\setlength{\tabcolsep}{4pt}
\begin{tabular}{l|cc}
\toprule
{Distortion Type} & {First Stage} & {Second Stage} \\
\midrule
None (Clean) & 0.700 & \textbf{0.983} \\
Resize (0.3×) & 0.557 & \textbf{0.679} \\
JPEG (50) & 0.575 & \textbf{0.723} \\
Crop (0.3×) & 0.695 & \textbf{0.970} \\
Rotation (25°) & 0.558 & \textbf{0.630} \\
Rotation (90°) & 0.479 & \textbf{0.483} \\
Brightness (×2) & 0.686 & \textbf{0.965} \\
Contrast (×2) & 0.573 & \textbf{0.757} \\
Saturation (×2) & 0.692 & \textbf{0.967} \\
Sharpness (×2) & 0.698 & \textbf{0.972} \\
Gaussian Noise & 0.619 & \textbf{0.882} \\
MPEG4 Compression & 0.590 & \textbf{0.723} \\
\bottomrule
\end{tabular}
\end{table}

Table~\ref{tab:second_stage} shows that the second stage of fine-tuning improves robustness for almost every distortion type.  
The first stage on static images helps the model learn spatial watermark features but does not expose it to motion or temporal correlations.  
Training on videos in the second stage adapts the decoder to video statistics and stabilizes message recovery across frames.  
The improvement is most evident under crop, brightness, and sharpness changes, where accuracy increases by more than 25 percent.  
Overall bit accuracy rises from 0.70 to 0.98 on clean data and improves under compression, noise, and geometric transformations.  
These results confirm that temporal adaptation strengthens watermark recovery and stability in realistic video conditions.

\subsection{Tamper Localization Threshold}
\label{sec:appendix_tamper}

This section provides additional details about the temporal tampering experiments and the similarity threshold analysis used in Section~\ref{sec:tamper_localization}.

\paragraph{Tampering Setup.}
We evaluate three common types of temporal manipulations that may occur during malicious video editing: 
\textit{(i) Frame Swap}, where two frames exchange positions to disrupt temporal order; 
\textit{(ii) Frame Insert}, where synthetic or unrelated frames are injected to mimic content insertion attacks; and 
\textit{(iii) Frame Drop}, where one or more frames are removed to simulate packet loss or targeted deletion. 
We also evaluate combinations of these operations to approximate realistic adversarial scenarios. For each manipulation type, tampered frames are generated randomly at different positions within the sequences of video frames used in our experiments.

\paragraph{Similarity Threshold Analysis.}
During verification, each extracted frame message is compared with the reference keys using Hamming similarity. A frame is flagged as inserted when its similarity to all reference keys falls below a given threshold $\tau$; otherwise, it is assigned to the closest match. 
We examine threshold values from 0.7 to 0.9, as reported in Table~\ref{tab:threshold_study}, to assess the trade-off between detection precision and recall. Higher thresholds produce stricter alignment but risk false positives, whereas lower thresholds tolerate distortions but may miss subtle manipulations. 
As shown in Table~\ref{tab:threshold_study}, $\tau = 0.8$ achieves the best overall balance, yielding more than 95\% localization accuracy across all single and combined tampering types.

\begin{table}[ht]
\centering
\footnotesize
\caption{Tamper Localization Accuracy at Varying Thresholds. Accuracy for detecting frame-level manipulations (Swap, Insert, Drop, and combinations) under different similarity thresholds. A threshold of 0.8 provides the best overall accuracy across all attack types.}
\renewcommand{\arraystretch}{1.3}
\label{tab:threshold_study}
\centering
\resizebox{0.5\textwidth}{!}{
\setlength\tabcolsep{2pt}
\begin{tabular}{c|ccccccc} 
    \toprule
    
    \multirow{2}{*}{Threshold}&\multicolumn{7}{c}{Accuracy $\uparrow$}\\ 
    
      & Swap & Insert & Drop & Swap \& Insert & Swap \& Drop & Insert \& Drop & Swap \& Insert \& Drop\\
    
    \midrule
    
    0.9& 0.855& 0.864 & 0.851& 0.864 & 0.855 & 0.861 & 0.862\\

    0.85& 0.936& 0.940 & 0.936 & 0.939 & 0.936 & 0.939 & 0.941\\

    \cellcolor{gray!30}{0.8} &\cellcolor{gray!30}{0.959} & \cellcolor{gray!30}{\textbf{0.962}}&\cellcolor{gray!30}{0.959}& \cellcolor{gray!30}{\textbf{0.960}} & \cellcolor{gray!30}{0.960} & \cellcolor{gray!30}{\textbf{0.962}} & \cellcolor{gray!30}{\textbf{0.959}} \\

    0.75&0.980 & 0.924 & 0.980 & 0.922 & 0.980 & 0.920 & 0.919 \\

    0.7& \textbf{0.989} & 0.931 & \textbf{0.989} & 0.931 & \textbf{0.988} & 0.927 & 0.927\\

    \bottomrule
    \end{tabular}
}
\end{table}

\subsection{\textbf{Impact of More Aggressive Tamper Localization Attacks}}
\label{sec:appendix_tamper_agressive}

To further evaluate the robustness of \sys's temporal tamper localization capabilities, we simulate increasingly aggressive attack scenarios by scaling the intensity of three canonical manipulations: \textit{frame swapping}, \textit{frame dropping}, and \textit{frame insertion}. While previous experiments tested single-instance manipulations, this ablation examines the system's tolerance under heavier tampering.

\begin{figure}
    \begin{center}    \centerline{\includegraphics[width=0.4\textwidth]{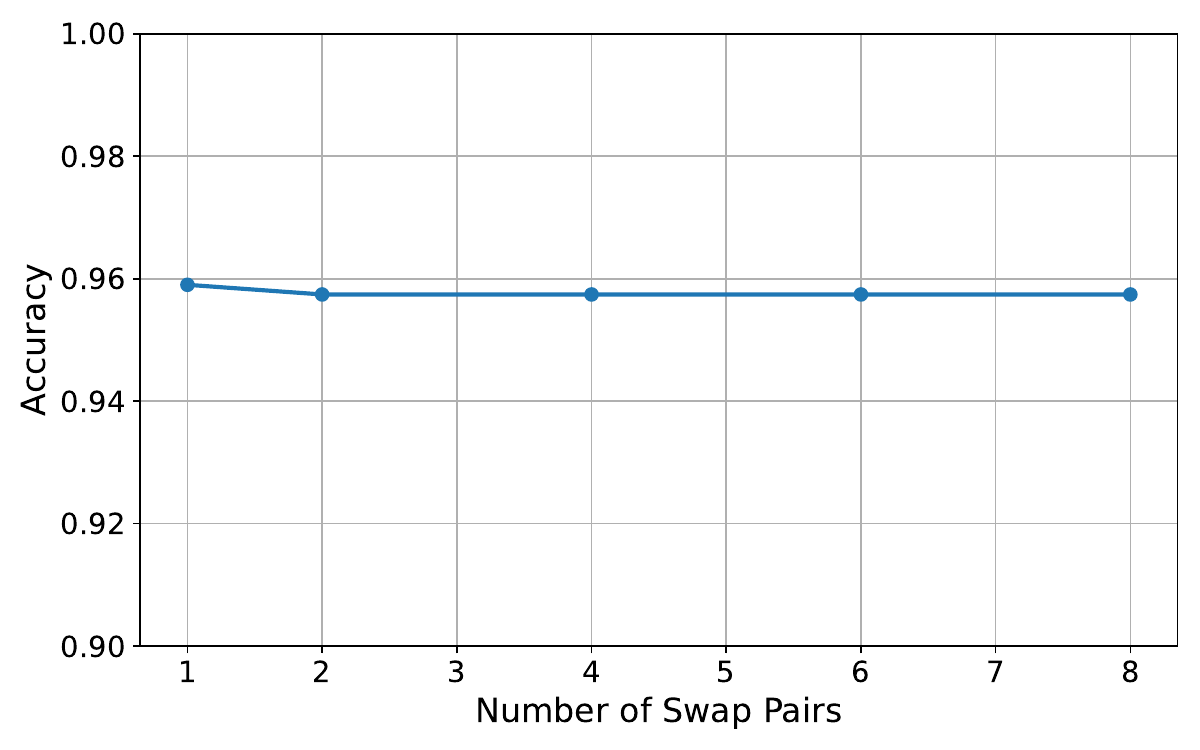}}
    \caption{Tamper Localization Accuracy vs. Number of Frame Swaps. Localization accuracy remains high even as the number of swapped frame pairs increases, showing \sys’s robustness to temporal reordering.}
    \label{fig:swap_pairs}
    \end{center}
\end{figure}

\paragraph{Swap Pairs} In this scenario, we randomly select and swap $N$ pairs of frames within each video. As shown in Figure~\ref{fig:swap_pairs}, localization accuracy remains consistently high as the number of swaps increases from 1 to 8. This result demonstrates \sys's robustness to moderate-to-severe temporal reordering.

\begin{figure}
    \begin{center}    \centerline{\includegraphics[width=0.4\textwidth]{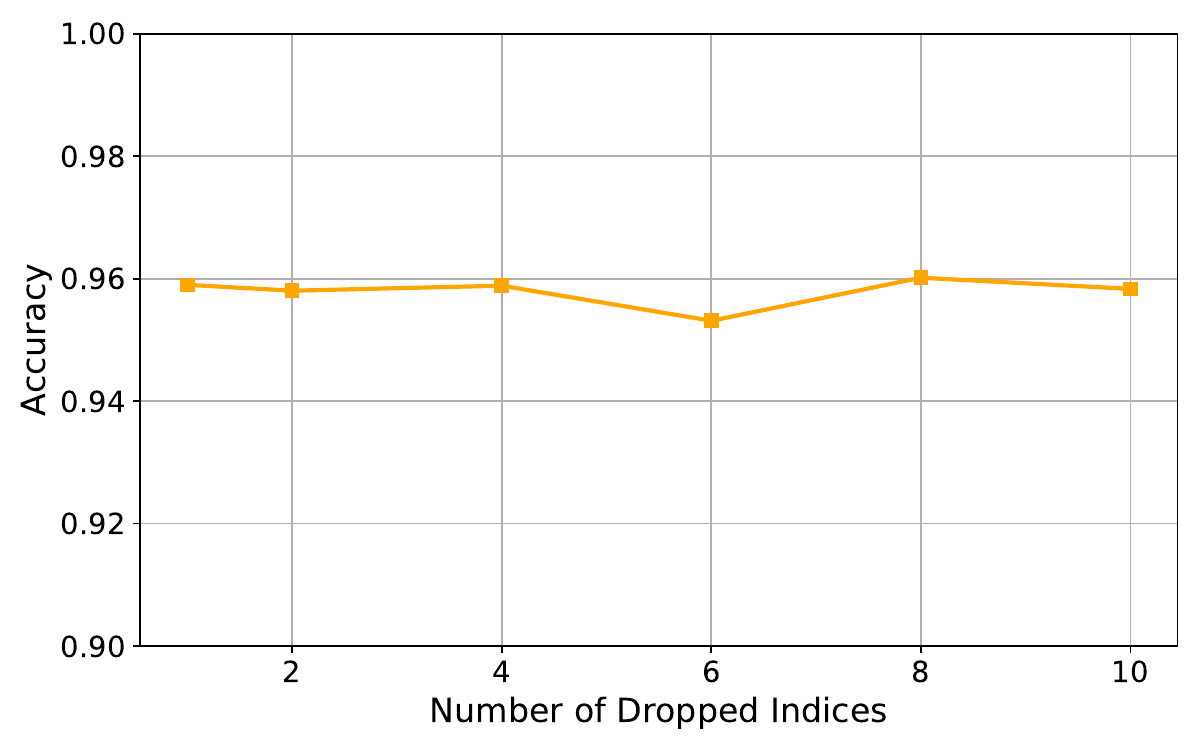}}
    \caption{Tamper Localization Accuracy vs. Number of Dropped Frames. \sys maintains strong localization performance despite the removal of multiple frames, demonstrating tolerance to frame loss.}
    \label{fig:drop_indices}
    \end{center}
\end{figure}

\paragraph{Drop Indices} We next evaluate performance under increasing numbers of dropped frames. As seen in Figure~\ref{fig:drop_indices}, localization accuracy degrades only slightly, even when up to 10 frames are missing. This suggests that the frame-message matching remains reliable despite partial sequence loss.

\begin{figure}
    \begin{center}    \centerline{\includegraphics[width=0.4\textwidth]{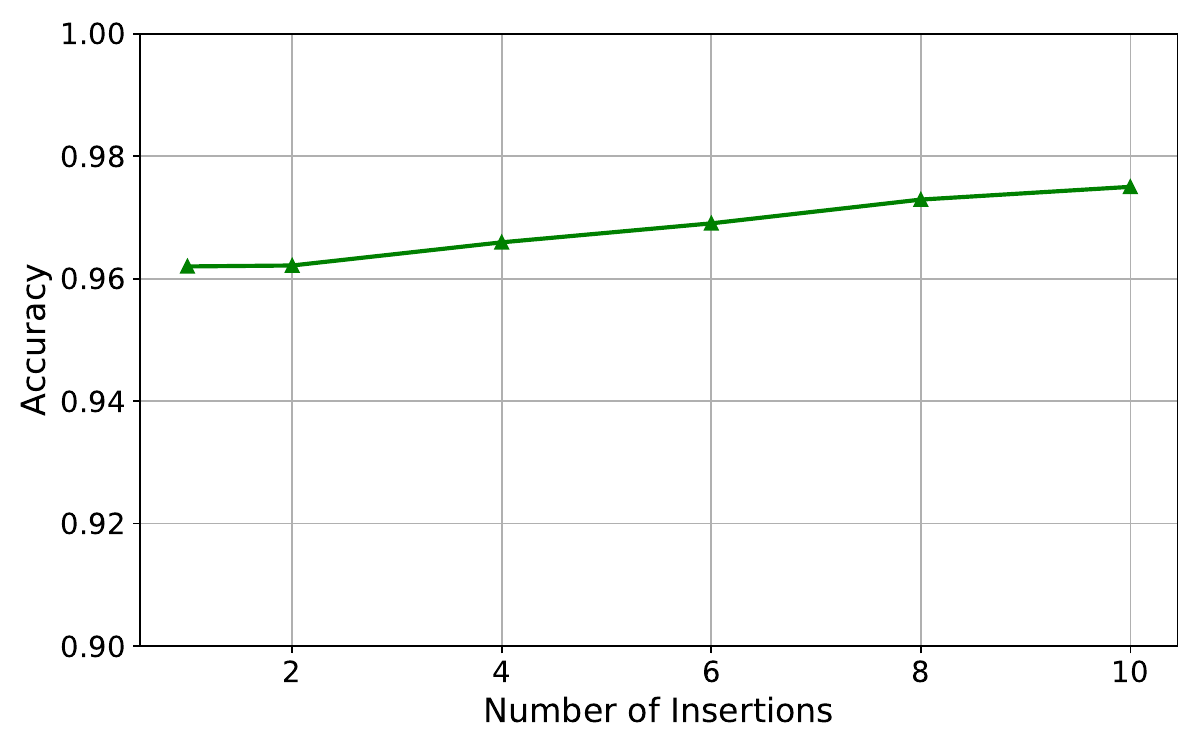}}
    \caption{Tamper Localization Accuracy vs. Number of Inserted Frames. Even with up to 10 random noise insertions, \sys accurately detects tampered frames, confirming its resilience to synthetic content injection}
    \label{fig:insertions}
    \end{center}
\end{figure}

\paragraph{Insertions} Lastly, we assess robustness to frame insertions by injecting randomly generated noise frames at random positions. Since the inserted frames are entirely unstructured noise and bear no resemblance to authentic watermarked content, our frame-wise message matching algorithm is able to detect them with \emph{very high accuracy}. As shown in Figure~\ref{fig:insertions}, detection accuracy remains high even when up to 10 synthetic frames are added. Interestingly, we observe a slight \emph{increase in overall localization accuracy} as the number of inserted noise frames grows. This is because the inserted frames are highly dissimilar to any of the template keys and are thus reliably flagged as tampered, effectively boosting overall detection performance.

\begin{figure}
    \begin{center}    \centerline{\includegraphics[width=0.4\textwidth]{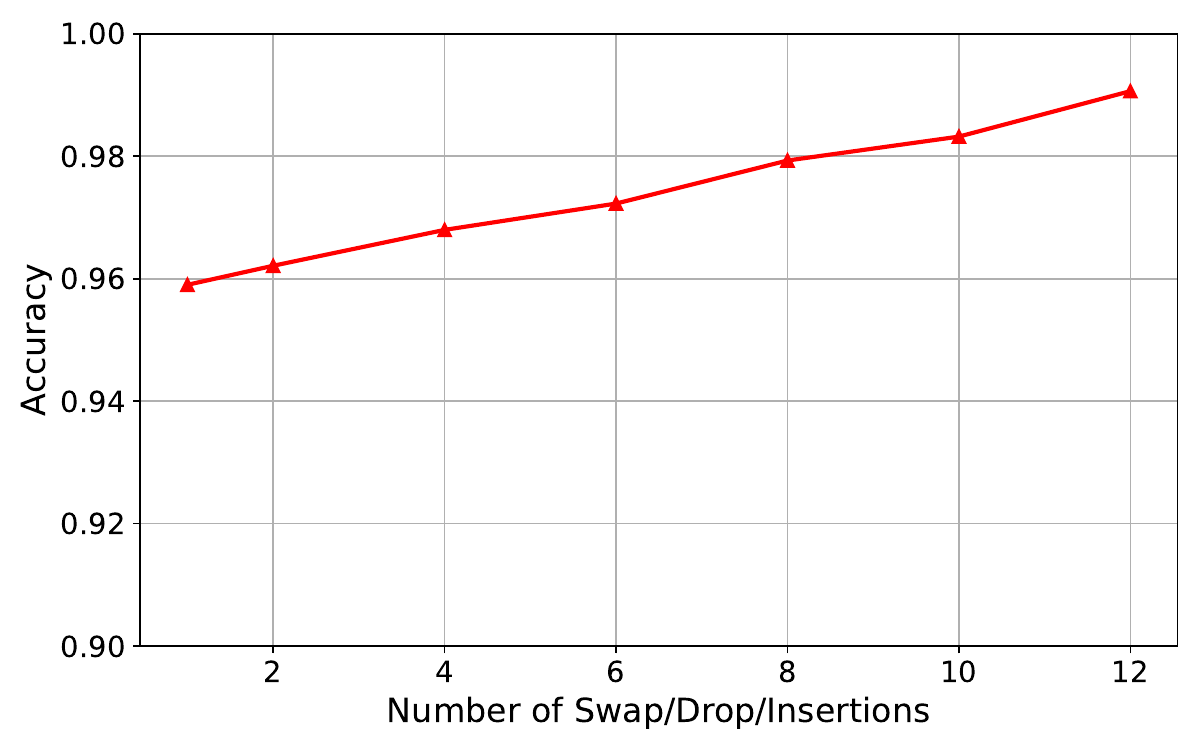}}
    \caption{Tamper Localization Accuracy vs. Number of Combined Attacks. Localization accuracy under simultaneous application of frame swaps, drops, and insertions. As the number of combined manipulations increases, accuracy remains high and exhibits a slight upward trend. This result reflects the cumulative behavior observed in Figures \ref{fig:swap_pairs}, \ref{fig:drop_indices}, \ref{fig:insertions} where swap and drop manipulations have minimal effect on accuracy, and insertions using random noise frames are consistently and easily detected—leading to an overall increase in detection performance as attack intensity grows.}
    \label{fig:overall}
    \end{center}
\end{figure}

\paragraph{Combined Attacks} In this experiment, we simultaneously apply all three types of tampering—frame swaps, frame drops, and insertions—by randomly selecting frames for each manipulation type and increasing their total number together. As shown in Figure~\ref{fig:overall}, localization accuracy remains consistently high and exhibits a slight \emph{upward trend} as the number of combined manipulations increases. This outcome aligns with our earlier observations: Figures~\ref{fig:swap_pairs},~\ref{fig:drop_indices}, and~\ref{fig:insertions} demonstrate that swap and drop attacks cause only a marginal decline in accuracy as they intensify, while insertion attacks using random noise frames actually lead to higher localization accuracy due to their detectability. As a result, when all three attack types are applied together, the increasing presence of easily identifiable inserted frames dominates the trend, yielding an overall increase in detection accuracy.

\noindent Overall, these results indicate that \sys remains effective under aggressive temporal tampering. Even when multiple manipulations are applied, the method consistently localizes frame-level alterations with high precision, supporting its applicability in real-world adversarial scenarios.

\begin{algorithm*}[ht]
\small
\caption{Temporal Tamper Localization}
\label{alg:tamper}
\begin{algorithmic}[1]
\Require Template keys $T \in \mathbb{R}^{M \times d}$, Tampered keys $K \in \mathbb{R}^{N \times d}$, True sequence $S \in \mathbb{Z}^N$, Threshold $\tau$
\Ensure Tamper localization accuracy
\State $P \gets$ empty list \Comment{Predicted frame sequence}
\For{$i = 1$ to $N$}
    \State $k_i \gets K[i]$ \Comment{Current decoded key}
    \State Compute Hamming similarity $sim$ between $k_i$ and all $T[j]$
    \State $j^* \gets \arg\max_j sim[j]$
    \If{$sim[j^*] < \tau$}
        \State $P.\text{append}(-1)$ \Comment{Inserted frame}
    \Else
        \State $P.\text{append}(j^*)$ \Comment{Best-matching original index}
    \EndIf
\EndFor
\State $accuracy \gets \frac{1}{N} \sum_{i=1}^{N} \mathbb{1}[P[i] = S[i]]$
\State \Return $accuracy$
\end{algorithmic}
\end{algorithm*}

\begin{figure*}    
    \begin{center}    \centerline{\includegraphics[width=1\textwidth]{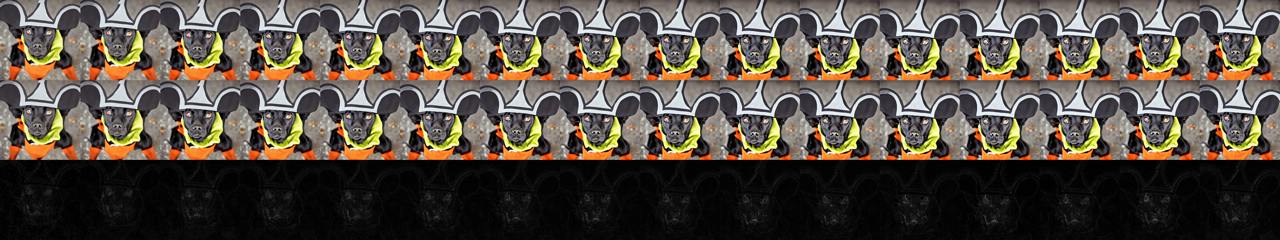}}
    \caption{Visual comparison between watermarked and non-watermarked outputs. The first row shows frames generated by \sys with embedded watermark messages. The second row shows the same frames generated without watermarking. The third row depicts the absolute pixel-wise difference between the two. The differences are visually negligible and imperceptible to the human eye. Most modifications are localized along object edges, where the model embeds message bits without introducing perceptible artifacts.}
    \label{fig:frames}
    \end{center}
\end{figure*}

\subsection{Visual Impact of Watermarking}

 In addition to quantitative evaluation, we also visualize the visual impact of our watermarking approach in Figure~\ref{fig:frames}, which shows sample frames from a generated video. The first row contains frames produced by \sys with embedded watermarks, while the second row shows frames generated by the original Stable Video Diffusion model without watermarking. The third row presents the absolute pixel-wise difference between the two outputs. As evident in the visualizations, the difference between the watermarked and non-watermarked frames is \emph{imperceptible to the human eye}; most changes are \emph{subtle and localized}.

Interestingly, the model appears to embed the watermark by modulating pixel values along the \emph{edges and contours of prominent objects} in the scene. This behavior is likely due to the \emph{higher spatial frequency content} in these regions, which allows information to be encoded without disrupting perceptual quality. The near-invisibility of the changes—even under frame-wise differencing—\emph{highlights the effectiveness of our perceptual loss formulation} and confirms that \sys successfully embeds information \emph{without introducing observable artifacts}.